\documentclass{new_tlp}

\let\proof\relax

\usepackage{amsmath}
\usepackage{graphicx}
\usepackage{multirow}

\usepackage{times}
\usepackage{soul}
\usepackage{url}
\usepackage[hidelinks]{hyperref}
\usepackage[utf8]{inputenc}
\usepackage[small]{caption}
\usepackage{graphicx}
\usepackage{amsthm}
\usepackage{booktabs}
\usepackage{algorithm}
\usepackage{algorithmic}
\usepackage[switch]{lineno}
\usepackage{enumitem}
\usepackage{amssymb}

\usepackage{todonotes}
\usepackage{listing}
 
\usepackage{booktabs}

\def\naf{{\; not \;}}
\usepackage{newfloat}
\usepackage{listings}
\DeclareCaptionStyle{ruled}{labelfont=normalfont,labelsep=colon,strut=off} 
\floatstyle{ruled}
\newfloat{listing}{tb}{lst}{}
\floatname{listing}{Listing}

\usepackage{color}
\usepackage{hyperref}
\usepackage{url}
\usepackage{comment}

\allowdisplaybreaks
\newtheorem{example}{Example}

\newtheorem{definition}{Definition}

\def\METHOD{ASPER}

\begin{document}

\title[ASP Enhanced NN Models for Joint Entity-Relation Extraction]
{\METHOD: Answer Set Programming Enhanced Neural Network Models for Joint Entity-Relation Extraction\thanks{Research partially funded by NSF awards \#1757207,  \#1914635, and \#1812628.}}

\author[T. H. Le, H. Cao, T. C. Son]
         {TRUNG HOANG LE, HUIPING CAO, TRAN CAO SON\\
         Department of Computer Science, New Mexico State University, Las Cruces, New Mexico, USA\\
         \email{\{trungle,hcao\}@nmsu.edu} and \email{tson@cs.nmsu.edu}}

\submitted{5 February 2023} \revised{8 April 2003} \accepted{11 May 2023}

\maketitle

\begin{abstract}
A plethora of approaches have been proposed for joint entity-relation (ER) extraction. Most of these methods largely depend on a large amount of manually annotated training data. 
However, manual data annotation is time consuming, labor intensive, and error prone. Human beings learn using both data (through  induction) and knowledge (through deduction). Answer Set Programming (ASP) has been a widely utilized approach for knowledge representation and reasoning that is elaboration tolerant and adept at reasoning with incomplete information. This paper proposes a new approach, \underline{ASP}-enhanced \underline{E}ntity-\underline{R}elation extraction (\METHOD), to jointly recognize  entities and relations by learning from both data and domain knowledge. In particular, \METHOD~takes advantage of the factual knowledge (represented as facts in ASP) and derived knowledge (represented as rules in ASP) in the learning process of neural network models. We have conducted experiments on two real datasets and compare our method with three baselines. The results show that our \METHOD~model consistently outperforms the baselines. 
\end{abstract}


\section{Introduction}
Entity-relation (ER) extraction is to identify named entities and relations from unstructured text. For joint ER extraction, deep neural network (NN) models have created many successful stories~(e.g., the papers by \citeN{DBLP:conf/coling/GuptaSA16}; \citeN{SPERT2020}; \citeN{wang2020two}). Despite such success, the supervised NN methods depend on utilizing a large amount of well-labeled training data. However, labeling free text data with entities/relations is time-consuming,  labor intensive, and error prone because of a lot of noise, as shown by \citeN{Chen_mingcai_2022_CoRR}.

Semi-supervised learning (SSL), introduced by \citeN{semi_supervised_learning_book_2006} and \citeN{semi_supervised_learning_2020}, has been utilized to improve predictions by using a small amount of labeled data and a much larger amount of unlabeled data. Among the many SSL approaches, the proxy-label methods described in the papers by \citeN{Ruder2018} and \citeN{semi_supervised_learning_2020} are one commonly utilized strategy. These approaches create different strategies to utilize the pseudo labels that are predicted from the unlabeled data. However, most of them do not make use of domain knowledge as discussed by \citeN{ACL2016_Hu}, which is abundant and very useful, in symbolic forms in the learning process as shown in the survey by \citeN{semi_supervised_learning_2020}. 

Recent years have witnessed the increasing interest in utilizing general domain knowledge (e.g., represented as symbolic knowledge) to improve machine learning models to alleviate the issue caused by the lack of large amounts of labeled data. Such efforts include neural symbolic modeling and abductive learning. 

Neural symbolic models, referred by us as works that encode the knowledge and rules to be a holistic component of neural network models (e.g., through the design of a new loss function).  
Such models have achieved great success (e.g.,~see the papers by \citeN{ACL2016_Hu}; \citeN{DBLP:journals/flap/GarcezGLSST19}). 
However, tightly modeling the symbolic knowledge as a part of NN models suffers from the elaboration tolerant issue where the model is hard to scale to changes of logical representations of facts  (e.g., loss functions need to be modified when  adding new rules).

\citeN{abductive_learning_Zhou2019}; \citeN{abductive_learning_neurips2019_Dai}; and \citeN{DBLP:conf/ijcai/CaiDHLM021}
introduced abductive learning that combines machine learning models (which are mainly data driven) and logic programming (which encodes background knowledge and reason about them). 
In abductive learning, an initial machine learning model $M$ is trained from the labeled data and used to get predicted labels from the unlabeled data (denoted as pseudo labels).
Pseudo labels may be wrong or inconsistent when the model $M$ is not effective. Example~\ref{eg_inconsistent_hidden_pseudo_labels} demonstrates different issues of pseudo labels. The pseudo labels are revised to get a new set of consistent pseudo labels through minimizing the inconsistency of the abduced labels and the knowledge. The revised set of pseudo labels are  used to retrain a machine learning model. 
Most existing abductive learning approaches use first order logic (FOL) to encode knowledge.    

In this work, we propose to design an SSL approach by encoding domain knowledge and rules using Answer Set Programming (ASP) for the joint recognition of entities and relations. The purpose of using the domain knowledge is to generate consistent pseudo labels (consistent w.r.t. the knowledge base) and derive more pseudo labels that cannot be predicted using the pure data driven models. ASP instead of FOL is used because of multiple advantages ASP provides.
ASP is a simple, rule-based, and declarative language that possess several theoretical building block results which support the development of provably correct programs. In addition, ASP is non-monotonic, which is important for dealing with commonsense knowledge, and supports an elaboration tolerant development of programs. For non-logical experts, ASP-rules are easier to use and to understand than FOL-formulae. 

The main contributions of this work are as follows. 
\begin{itemize}
    \item A new framework, \underline{ASP}-enhanced \underline{E}ntity-\underline{R}elation extraction (\METHOD), is introduced to make use of sophisticated domain knowledge in neural network models. ASP-encoded knowledge and rules intend to generate higher quality pseudo labels, which are further used to improve the model. As far as we know, this is the {\bf first work} that incorporates logic programming to deep learning models for joint ER extraction.
    \item \METHOD~introduces novel commonsense rules to select pseudo labels that may improve the model performance with higher probabilities. 
    \item The experimental evaluation on two real datasets shows that the proposed \METHOD~consistently improves the other baselines.
\end{itemize}
In what follows, Section~\ref{sec:relatedwork} reviews the related work, Section~\ref{sec:problem} formally defines the problem and related notations, Section~\ref{sec:method} explains our new framework, Section~\ref{sec:exp} shows our experimental results, and Section~\ref{sec:conclusion} concludes the work. 

\vspace{-0.15in}
\section{Related Work}
\label{sec:relatedwork} 
SSL methods are designed to make use of a small amount of labeled data and a much larger amount of unlabeled data in the learning process.
Traditional SSL methods including self-training and tri-training have been revisited recently by \citeN{Ruder2018}. 
Both self-training and tri-training utilize an iterative process to improve the initial model(s) trained on the small amount of labeled data by iteratively adding pseudo labels to the training set. 
In each iteration, self-training picks pseudo labels that have higher prediction probability and tri-training picks the pseudo labels that are agreed by at least two models.
A surprising finding in such revisit is that the classic tri-training, introduced in the paper by \citeN{tri-training2005}, strategy with minor changes can outperform many recent NN models. 
 
Many recent SSL approaches have been proposed to conduct ER extraction. \citeN{DBLP:conf/emnlp/HuZMLWY21} proposes a self-training based SSL method for relation extraction. In each iteration, it adopts meta-learning to generate higher quality pseudo labels. 
Curriculum labeling, introduced in the paper by \citeN{curriculum_labeling_aaai2021}, borrows the idea of curriculum learning discussed in the paper by \citeN{curriculum_learning_Bengio_2009}, which uses easy samples first and proceeds to use hard samples, and proposes a self-paced curriculum (curriculum labeling) in the pseudo-labeling process. However, this model does not use any symbolic knowledge. Most of these approaches do not make use of domain knowledge in symbolic forms.

Similar to SSL, neural network models also alleviate the issue of insufficient labeled data. \citeN{ACL2016_Hu} is the first work of integrating logic rules with NN models. It proposes an iterative distillation method to enhance several NNs with declarative first-order logic (FOL) rules, which is encoded using soft logic. 
NeurASP, introduced in the paper by \citeN{NeurASP_yang_ijcai2020}, also employs ASP and neural networks. Its knowledge base is predefined and all the atoms are utilized in the learning process. However, our knowledge base is used to generate answer sets with consistent pseudo labels and some answer sets (when multiple are available) may not be utilized in the learning process. 

Our work also shares similarity with the framework of abductive learning such as those described in the papers \citeN{abductive_learning_Zhou2019}; \citeN{abductive_learning_zhou_2020} in that symbolic knowledge is used to improve the quality of pseudo labels.  
However, our work is different from abductive learning in several aspects. Abductive learning (e.g.,~the paper by \citeN{abductive_learning_zhou_2020}) derives abduced pseudo labels through an optimization process. Once these labels are revised, they are used to retrain a model. 
When they retrain a model, all the pseudo labels are utilized.
Our approach utilizes a subset of the pseudo labels, which have higher probability to be true, to retrain a model. In addition, the pseudo labels are iteratively refined using ASP, which provides powerful reasoning capability.

\vspace{-0.15in}
\section{Problem Definition and Terminology}
\label{sec:problem}
This section defines our research problem and related terminology.   

\smallskip \noindent
{\bf Problem definition}: Given a dataset $D_L$ (training data) with labeled entities and relations, a dataset $D_{UL}$ without any annotated labels, where $|D_L| \ll |D_{UL}|$, and a knowledge base\footnote{
The KB is domain-dependent. We discuss practical ideas on developing such KB for the joint ER extraction problem in the later section. 
} $KB$ which encodes the common sense facts and reasoning rules, our problem is to learn an NN model $M \leftarrow f(D_L, D_{UL}, KB)$ to capture the hidden patterns about entities and relations in the datasets $D_L \cup D_{UL}$.  

\begin{definition}[Pseudo labels]
Given a model $M$ and a dataset $D_{UL}$ without any annotated labels, the predicted labels from $D_{UL}$ using $M$ are called pseudo labels. 
\end{definition}

The possible entity and relation labels that occur in the training data $D_L$ are represented as $ent$ and $rel$. Given any token or word in a sentence, we use $b$ and $e$ to represent the beginning and ending locations of that token in the sentence. The $b$ and $e$ are in the range of 0 and the total number of tokens of a sentence.
An entity pseudo label is in the form of 
\begin{eqnarray}
ent(b, e), conf   \label{entity_pl} 
\end{eqnarray}
It means that the tokens at the locations [$b$, $e\!-\!1$] in the sentence is of entity type $ent$. 
Here, $conf$ is a value in the range of [0,1] indicating the confidence of the prediction.\\
A relation pseudo label is in the form of 
\begin{eqnarray}
rel(b, e, b', e'), conf  \label{relation-pl}
\end{eqnarray}
\noindent Here, $b$ ($b'$) and $e$ ($e'$)  represent the beginning and ending locations of the first (second) token in the relation. To make the descriptions more intuitive, we sometimes represent  a relation as 
\begin{eqnarray}
rel(tokens, tokens'), conf \label{relation-pl2}
\end{eqnarray}
where $tokens$ ($tokens'$) are the first (second) tokens at locations [$b$, $e$) (and [$b'$, $e'$)).

\noindent Without loss of generality, we may omit $conf$ when writing the entity and relation pseudo labels.

\begin{example}[Running example notations]
In later examples, we will use some well defined entity types. For example, $org$, $loc$, and $peop$ represent organization, location, and people respectively. 
Some predefined relations are $livedIn$, $locatedIn$, and $orgBasedIn$. They describe one person, a location, and an organization lives in, is located in, or is based in a location respectively.
\end{example}

\vspace{-0.15in}
\section{ASP Enhanced Neural Network Models for  Entity-Relation Extraction (\METHOD)}
\label{sec:method}

This section presents our proposed \METHOD~method. \METHOD~targets at utilizing answer sets and ASP to improve the quality of pseudo labels and derive more useful labels to retrain the model. 

\begin{algorithm}[b]
\caption{\METHOD~framework}
\label{alg:algorithm}
\begin{footnotesize}
\textbf{Input}: Labeled data $D_L$; Unlabeled data $D_{UL}$; Knowledge Base $KB$\\
\textbf{Parameter}: Confidence step parameter $\Delta$ (e.g., 20)\\
\textbf{Output}: The model $M$
\begin{algorithmic}[1]
\STATE Learn an initial model $M$ from $D_L$
\label{step_initialmodel} 
\STATE $\Delta_t = 100 - \Delta$
\WHILE{iteration condition is not met}\label{step:begin_while}
    \STATE $\mathbb{Z} \leftarrow \emptyset$ \# {\em The set of the selected answer sets for all the sentences in $D_{UL}$}
    \STATE $D_{aug} \leftarrow \emptyset$ \# {\em The pseudo labels augmented to train the model}
    \FOR{each sentence $x \in D_{UL}$}
        \STATE $z$ $\leftarrow$ GenPseudoLabels($M$, $x$)
        \label{step:generate_pl}
        \STATE $A_S \leftarrow$ {\bf Convert2Atoms}($z$) 
        \label{step:pl_to_atoms}
        \STATE $W$ $\leftarrow$ {\bf ReviseUsingASP}($A_S \cup KB$) 
        \#{\em $W$ is an answer set associated with a confidence value $W.conf$}
        \label{step:asp}
        \STATE $\mathbb{Z} \leftarrow \mathbb{Z} \cup W$
        \label{step:upd_atoms}
    \ENDFOR
    \STATE \label{step:cal_conf_val} 
    $T \leftarrow $ the confidence value at $\Delta_t$ percentile of all the answer sets in $\mathbb{Z}$
    \label{step:sel_sentences}
    \FOR {each answer set $W \in \mathbb{Z}$}
        \IF{$W.conf \ge T$}
            \STATE $D_{aug} \leftarrow D_{aug} \cup W$
        \ENDIF
    \ENDFOR
    \STATE \label{step:retrain_model} Train model $M$ on $D_L \cup D_{aug}$  from scratch
    \STATE \label{step:reduce_deltat}
    $\Delta_t = \Delta_t - \Delta$
\ENDWHILE \label{step:end_while}
\STATE \textbf{return} $M$
\end{algorithmic}
\end{footnotesize}
\end{algorithm}

\vspace{-0.15in}
\subsection{\METHOD~Framework}
The framework of \METHOD~is shown in Algorithm~\ref{alg:algorithm}. \METHOD~first trains an initial model using the limited amount of training data (Line~\ref{step_initialmodel}) and improves the model through an iterative process using ASP revised pseudo labels (Lines~\ref{step:begin_while}-\ref{step:end_while}).
 
To train an initial neural network model, we utilize the SpERT architecture proposed in the paper by \citeN{SPERT2020} due to its lightweight nature. 
Multiple iterations (Steps~\ref{step:begin_while}-\ref{step:end_while}) are used to improve the model. In each iteration, \METHOD~predicts the entities and relations in a sentence $x$ (i.e., the pseudo labels)  using the model $M$ (Line~\ref{step:generate_pl}) where $M$ can be the initial model (trained in Line~\ref{step_initialmodel}) or the retrained model (Line~\ref{step:retrain_model}).
Then, it utilizes ASP to update these pseudo labels (Lines~\ref{step:pl_to_atoms}-\ref{step:upd_atoms}). The updated pseudo labels coupled with the selected sentences are then used to retrain the model (Lines~\ref{step:sel_sentences}-\ref{step:retrain_model}). There are many ways to revise pseudo labels. We define a preference relation over the sets of revised labels (answer sets) based on the notion of the probability of a set of revised labels (Definition~\ref{def:answerset_prob} in Section~\ref{sec:asp}). This preference relation is then used to select the most preferred set of revised labels. The iteration condition can be that the prediction of the unlabeled data does not change anymore or the number of iterations reaches a threshold. In our experiments, we set the number of iterations to be a fixed number. 

\vspace{-0.15in}
\subsubsection{Generate pseudo labels} 
Using the initially trained model or a model that is trained in the previous iteration $M$, the algorithm can recognize the entities and relations in the unlabeled dataset $D_{UL}$. The predicted entity and relation pseudo labels are in the form of Eqs.~\eqref{entity_pl} and \eqref{relation-pl}. 

\begin{example}[Pseudo labels] \label{eg:pl}
Given a sentence ``CDT Tuesday is in the area of Port Arther and Galveston, Texas.'' the predicted pseudo labels look like the following:

\noindent
\begin{tabular}{|lr|lr|lr|}
\cline{1-6} org(0,2), & 0.888 & other(1,2), & 0.799 &  locatedIn(7,9,10,11), &  0.998    \\
loc(7,9), & 0.998 & loc(10,11), &  0.998 & locatedIn(0,2,12, 13), &  0.993  \\
loc(12,13), &  0.998 & orgBasedIn(1,2,12,13), & 0.777 & locatedIn(10,11,12,13), & 0.993  \\\cline{1-6}
\end{tabular}
The pseudo label ``org(0,2), 0.888'' means that the token from location 0 to 1 (which is ``CDT Tuesday'') is an organization ($org$). This prediction has confidence value 0.888. Similarly, the pseudo label ``loc(7,9)'' means that ``Port Arther'' (the tokens at locations 7 and 8) is of type location ($loc$). 
Correspondingly, the predicted relation ``locatedIn(7,9,10,11)'' means that ``Port Arther'' is located in ``Galveston''. The other entities and relations can be interpreted accordingly. To assist the understanding of the relations, we create a figure for these entities and relations. 
\begin{figure}[htb]
    \centering
    \includegraphics[width={.99\columnwidth}]{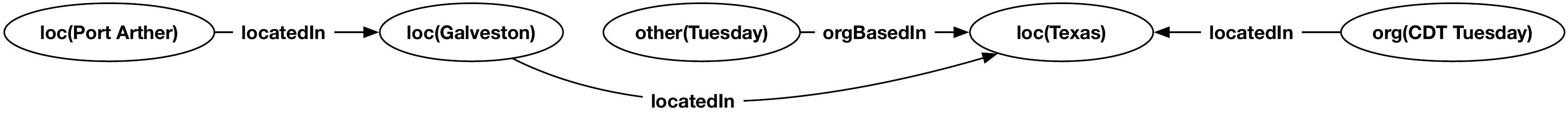}
    \caption{Example of pseudo labels (a node represents an entity and a directed edge between a source (first tokens) and a destination (second tokens) represents a relation)}
    \label{fig:example}
\end{figure}
\end{example}

\vspace{-0.15in}
\subsubsection{Improve pseudo-label quality using answer sets}
The predicted entity and relation pseudo labels may be wrong (just like any machine learning model does) or inconsistent.

\begin{example}[Inconsistent and hidden pseudo labels]
\label{eg_inconsistent_hidden_pseudo_labels}
Example~\ref{eg:pl} shows the pseudo labels predicted from one sentence. They have different types of inconsistencies. 
\begin{list}{$\bullet$}{\topsep=1pt \parsep=0pt \itemsep=1pt \leftmargin=10pt} 
\item (inconsistent labels) The relation $locatedIn({CDT~Tuesday}, Texas)$ and entity $org({CDT~Tuesday})$ are not consistent because the first term of $locatedIn$ needs to be a location, but $CDT~Tuesday$ is an organization. Similarly, the entity $other(Tuesday)$ and relation $orgBasedIn(Tuesday, Texas)$ are not consistent because the first term of $orgBasedIn$ needs to be an organization. 
\item (overlapping entities) The two entities $org(0,2)$ and $other(1,2)$ overlap because ``Tuesday'' is a part of ``CDT Tuesday''. It makes more sense not to have both. 
\item (hidden labels) Given $locatedIn({Port~ Arther}, Galveston)$ and $locatedIn(Galveston, \newline Texas)$, another relation  $locatedIn(Port~Arther, Texas)$ should be valid. 
However, the model does not predict this relation. 
\end{list}
\end{example}

In utilizing pseudo labels to improve a model, a recognized problem is {\em confirmation bias} as mentioned in the paper by \citeN{curriculum_labeling_aaai2021}, which means that utilizing the wrong pseudo labels to retrain a model can amplify the error. 
It is critical to control which pseudo labels are utilized to retrain the model.  

The issues listed above are not inclusive. Their root issue is the ineffective pure data-driven model learned from insufficient training data. The most commonly identified issue with an ineffective model is its inaccurate predictions. We target at addressing the root issue by somehow correcting the wrongly predicted pseudo labels.
Our \METHOD~framework (Lines~\ref{step:pl_to_atoms}-\ref{step:upd_atoms}) improves the quality of pseudo labels by computing a consistent set of pseudo labels (an answer set).
It first converts all the pseudo labels ($z$) to  a set of atoms $A_S$ that ASP can process (using Function {\em Convert2Atoms}). 
Given $A_S$ and the knowledge base $KB$, there may be multiple answer sets. {\em ReviseUsingASP} utilizes the rules in the $KB$ to calculate a probability for each answer set and chooses the one with the highest probability and associates with it a confidence level. The details of the two steps are described in Section~\ref{sec:choose_answer_set}. 

\vspace{-0.15in}
\subsubsection{Model retraining with improved pseudo labels}
\label{sec:model_retrain}

Once we get the improved pseudo labels from the  unlabeled dataset $D_{UL}$, these improved pseudo labels are put to $\mathbb{Z}$ (Line~\ref{step:upd_atoms}) and are used to help retrain the model. 

We observe that some answer sets have much higher confidence values than others. The pseudo labels in these answer sets tend to be correct with higher probabilities. Based on this observation, the model retraining first utilizes the pseudo labels in the answer sets with higher confidence values and proceeds to use pseudo labels in answer sets with lower confidence values. 
This idea is the same as that in curriculum labeling proposed by~\citeN{curriculum_labeling_aaai2021} that uses a portion (with high prediction confidence) of the pseudo labels to retrain a model in each iteration.  
This curriculum idea is implemented through the use of $\Delta_t$ in Line~\ref{step:cal_conf_val}. 
With the iterations proceed, $\Delta_t$ decreases (Line~\ref{step:reduce_deltat}). 
At the end, when $\Delta_t$ becomes zero, the model retraining uses the pseudo labels in all the answer sets. 

\vspace{-0.15in}
\subsection{Computing Improved Pseudo Labels via ASP}
\label{sec:asp}

\paragraph{Background.}
ASP, proposed in the papers by \citeN{MarekT99}; and \citeN{Niemela99}, is a knowledge representation and reasoning (KR\&R) approach to problem solving using logic programs under answer set semantics introduced by \citeN{GelfondL90}. In ASP, a problem is solved by first encoding it as an ASP program, whose answer sets correspond one-to-one to the solutions of the problem. 
The encoded ASP program is then given to an ASP solver (e.g., \lstinline{clingo}~as described in the paper by \citeN{gekakasc14b}) to compute answer sets and solutions can be extracted from these answer sets.

The ASP language also includes language-level extensions to simplify the use of ASP in practical applications. 
We will make use of the \emph{choice atom} of the form $l \: \{ l_1;\ldots;l_n \} \: u$ where 
$l_i$ is an atom, $l$ and $u$ are integers.  
Intuitively, it says that the number of literal $l_i$ that is true must be within a lower bound $l$ and an upper bound $u$. 
If $l$ (resp. $u$) is not specified, then the lower (resp. upper) bound is $0$ (resp. $+\infty$) by default. 
A choice atom can appear in the head or body of a rule or after the default negation.  

\subsubsection{ASP for computing pseudo labels}
First, the pseudo labels representing the entities (Eq.~\eqref{entity_pl}) and relations (Eq.~\eqref{relation-pl}) are represented in ASP using atoms of the form \eqref{entity} or \eqref{relation}. 

\begin{minipage}{.44\textwidth}
    \begin{eqnarray}
       atom(entity(ent, b, e), conf)    \label{entity} 
     \end{eqnarray}
\end{minipage}
\begin{minipage}{.55\textwidth}
    \begin{eqnarray}
   atom(relation(rel, b, e, b', e'), conf) \label{relation} 
     \end{eqnarray}
\end{minipage}

\noindent Let $A_S$ be the collection of atoms of the form \eqref{entity} or \eqref{relation} for a sentence $S$.
A sentence is a part of a dataset $D$ that often has declarative knowledge associated with it.  
In this paper, we consider the different types of knowledge that are usually available given $D$ which can be specified as follows: 

\smallskip \noindent 1.    
   \emph{Type declaration}: a type declaration defines the type of a relation and is given in the form 
   \begin{equation} \label{typedef} 
      type\_def(rel, ent, ent'). 
   \end{equation} 	
   A type declaration by Eq.~\eqref{typedef} says that relation $rel$ is between entities $ent$ and $ent'$. 
   For example, in the domain \lstinline{CoNLL04} (see next section), 
   the relation $liveIn$ is specified by the atom 
   $type\_def(liveIn, peop, loc)$ which says that it is a relationship between entities of the types $peop$ and $loc$. 
   
\smallskip \noindent 2.     \emph{Inference rule}: in many domains, there are relationships among relations. For example, $locatedIn$ is transitive in the sense that if area $A$ is located in area $B$ and $B$ is located in $C$ then $A$ is located in $C$. This rule can easily be specified by an ASP rule of the following form\footnote{ 
We use variables (strings starting with an uppercase letter) in the logic program. A rule with variables is the collection of ground rules obtained from substituting variables with any possible values; in this case, variables refer to locations in the sentence.}:  
   \begin{equation} \label{inf:rule} 
      rule(X, Y, Z) \leftarrow Body  
   \end{equation} 	
   where $X$, $Y$, and $Z$ is of the form $relation(R, B, E, B', E')$ and $Body$ is domain-specific information. 
   The rule relating to $locatedIn$ discussed above can be encoded as follows:  
    \begin{eqnarray*}
     & & rule(relation(locatedIn,B_1,E_1,B_2,E_2),  relation(orgbasedIn,B_o,E_o,B_1,E_1), \\
     & & \quad \quad \underline{relation(orgbasedIn,B_o,E_o,B_2,E_2)}) \leftarrow \\ 
     & & atom(relation(locatedIn,B_1,E_1,B_2,E_2)), atom(relation(orgbasedIn,B_o,E_o,B_1,E_1)), \\ 
     & &  \!\! \naf atom(relation(orgbasedIn,B_o,E_o,B_2,E_2)). 
    \end{eqnarray*}
    The head of the above ASP rule encodes an inference rule, whose first two labels (the relations on the first line) are predicted but the third relation (underlined) is missing in the set of predicted labels. 
    This inference rule is used for inferring the third pseudo label only if the first two pseudo 
    labels exist and the third pseudo label is not in the predicted model. 

\smallskip \noindent 3. 
    \emph{Optional parameters}: in some dataset, entity pseudo labels cannot overlap and/or each sentence has at least one relation. Such information can be specified by setting different flags. In our experiments, we use the following: 
    \begin{eqnarray}
        overlap\_fl.  && \quad \%\textnormal{ true if present} \label{flag:overlap} \\
       relation\_fl. && \quad \%\textnormal{ true if present} \label{flag:relation}
    \end{eqnarray}
     Form~\eqref{flag:overlap} forbids overlapping entities while \eqref{flag:relation} signals that each sentence should have a relation. 
     
\smallskip \noindent 4. 
    \emph{Other rules}: any ASP rule can be a part of the domain-dependent part. Preferably, these rules work with the predicates defined below. 

We refer to the collection of rules of the form \eqref{inf:rule}--\eqref{flag:relation} and any other rules for a domain $D$ as $KB_D$. 
We denote that $A_S$ is {\em inconsistent} when it contains pseudo labels that either contradict to each other or violate the rules in $KB_D$.

We next describe $\Pi$, the ASP program that takes the domain dependent knowledge $KB_D$ and the set of pseudo labels for a sentence $A_S$ as inputs and produces a consistent set of pseudo labels. 
$\Pi$ uses the following main predicates:
\begin{list}{$\bullet$}{\topsep=1pt \parsep=0pt \itemsep=1pt \leftmargin=10pt} 
    \item $pi(X)$: the prediction of $X$ (entity/relation pseudo label) might be incorrect ($pi$ stands for $possible\_incorrect$); 
    \item $ok(X)$: the prediction of $X$ is considered as correct; as such, $X$ can be used as pseudo label (for further training); 
    \item $nok(X)$: $X$ is not included in the output (set of pseudo labels); and
    \item $inf(X)$: $X$ is derived using 
    domain-dependent rules.
\end{list}

\noindent 1. \underline{\emph{Overlap checking rules}}: $\Pi$ contains the following rules:
\begin{eqnarray}
    && 2\{pi(entity(N, B, E)); pi(entity(N', B', E'))\}  \leftarrow   overlap\_fl, \nonumber \\
    && \quad \quad atom(entity(N, B, E)),  B < B', E > B',  atom(entity(N', B', E')).\label{over1} \\
    && \leftarrow overlap\_fl, ok(entity(N, B, E)),  B < B', E > B',   ok(entity(N', B', E')).\label{cover1} 
\end{eqnarray}
Rule~\eqref{over1} states that if the starting location $B'$ of an entity ($N'$) lies within the interval of the some other entity ($N$) then the prediction of the two entities might be wrong which is represented by the predicate $pi$. 
This leads to the constraints~\eqref{cover1} that prevents the consideration of both entities at the same time when they overlap in a sentence. The presence of $overlap\_fl$ in these rules indicates that they are in effect only if $overlap\_fl$ is set to true in $KB_D$. Similar rules and constraints are also implemented for the case that the starting indices of both entities are the same. They are omitted for brevity to save space and are listed in the appendix.

\smallskip \noindent 
2. \underline{\emph{Type checking rules}}: 
This type of rules is used to make sure that the entity types and relation types are consistent with the type declaration information in $KB_D$: 
\begin{eqnarray} 
&&\!\!\!\! 2\{pi(relation(R, B, E, B', E')); pi(entity(N, B, E))\} \leftarrow type\_def(R, N_1, N_2),\nonumber \\
&&  \quad \quad 	atom(relation(R, B, E, B', E')), 	atom(entity(N, B, E)),  N_1 \ne N. \label{dep1}  \\
&&\!\!\!\!\!  \leftarrow ok(relation(R, B, E, \_, \_)), ok(entity(N, B, E)), type\_def(R, N', \_), N \ne N'. \label{cdep1}  
\end{eqnarray} 
Rule~\eqref{dep1} states that if the first entity in a predicted relation is different from its specified type then both the predicted relation and the entity might be wrong.  
Constraint \eqref{cdep1} disallows the acceptance of both the relation and entity if their types do not march the specification. 
Again, we omit some rule relating to the second entity and the relation.

\smallskip \noindent 
3. \underline{\emph{Type inference rules}}: These rules use the type declarations from $KB_D$ to infer the type of accepted entities or the possible incorrect predictions of relations and entities. 
\begin{eqnarray}
&&\!\!\!\!\!\! 2\{ok(entity(N, B, E)); ok(entity(N', B', E'))\}  \leftarrow  \nonumber \\ 
&& \: type\_def(R, N, N'), ok(relation(R, B, E, B', E')). ~~~~\label{ok1} \\
&&\!\!\!\!\!\!  pi(entity(N', B', E'))   \leftarrow atom(relation(R, B, E, B', E')),   \nonumber \\ 
                    &         & \: pi(entity(N, B,E)), type\_def(R, N, N'). ~~\label{ok3} 
\end{eqnarray}
Rule~\eqref{ok1} says that if 
a relation $R$ is accepted as a true prediction  then 
we should also accept its first and second entity with the  type 
specified by the type declaration of $R$.  
Rule~\eqref{ok3} indicates that if the first entity of a relation $R$ is potentially incorrect then so is its second entity. 

\smallskip \noindent 
4. \underline{\emph{Rules for inference from predictions}}: The following rules are used for processing an inference rule specified in $KB_D$ via atom of the form $rule(X,Y,Z)$. 
    \begin{eqnarray}
      &&\!\!\!\!\!\!\!\!\!   6 \{ pi(X); pi(Y); pi(Z); inf(Z); dependency(X, Z); dependency(Y, Z) \} \leftarrow  \nonumber \\
      &&  \quad \quad  rule(X, Y, Z),  atom(X),   atom(Y), \naf  atom(Z).~~~~~ \label{ind10} \\ 
    &&\!\!\!\!\!\!\!\!\!  \leftarrow ok(Y), inf(Y), dependency(X,Y),not\: ok(X). ~~~~~~~~\label{ind11}\\
    &&\!\!\!\!\!\!\!\!\!  \leftarrow rule(X, Y, Z), ok(X), ok(Y), \naf ok(Z). ~~~\quad \label{ind12}
    \end{eqnarray}
Rule \eqref{ind10} says that if we have an inference rule $rule(X,Y,Z)$ and $X$ and $Y$ were predicted but not $Z$ then $Z$ is an inferred prediction and all the predictions might be incorrect. Furthermore, $Z$ depends on $X$ and $Y$. 
    Constraint \eqref{ind11} states that if $Y$ is an inferred atom and depends on $X$ then the acceptance of $Y$ cannot be done separately from the acceptance of $X$.   
    Constraint \eqref{ind12}, on the other hand, states that if $rule(X,Y,Z)$ is an inference rule then the acceptance of $X$ and $Y$ cannot be done separately from the acceptance of $Z$.   

\smallskip \noindent 
5. \underline{\emph{Rules for checking the existence of relation}}: The following rule records the acceptance of some relation pseudo label whenever $relation\_fl$ is defined: 
    \begin{eqnarray}
        relation\_exists  \leftarrow  ok(relation(R, \_,\_,\_,\_)), relation\_fl. \label{re1}   
    \end{eqnarray}
        
\smallskip \noindent
{6. \underline{\emph{Rules for selection of a consistent set of pseudo labels}}}: This set of rules defines the various types of atoms that will be used for computing the probability of a set of pseudo labels and selecting a set of pseudo labels. 

\begin{minipage}{.45\textwidth}
    \begin{eqnarray}
       atom(X)  \leftarrow  atom(X, P). \label{ind1} \\
        prob(X, P)  \leftarrow  atom(X, P), ok(X). \label{ind2} \\
        ok(X)   \leftarrow  atom(X), \naf pi(X). \label{ind4}
     \end{eqnarray}
\end{minipage}
\begin{minipage}{.55\textwidth}
    \begin{eqnarray}
   && 
   \!\!\!\!\!    invprob(X, P)   \leftarrow  atom(X, P), nok(X). \label{ind3} 
        \\
   && \!\!\!\!\!    \{ ok(X) \}   \leftarrow  1\{atom(X);inf(X)\}, pi(X). \label{ind5} \\
   && \!\!\!\!\!    nok(X)   \leftarrow  1\{atom(X);inf(X)\}, \naf ok(X). ~~~~~~~\label{ind7}
     \end{eqnarray}
\end{minipage} 
\\

\noindent Rule \eqref{ind1} projects an input $atom(X, P)$ to define $atom(X)$ for use with other part of the program.  
Rules~\eqref{ind2}--\eqref{ind3} define the predicates $prob$ and $invprob$ that are used in computing the probability of  the set of selected labels (see later). 
Rule \eqref{ind4} says that if there is no information about the potential incorrectness of the prediction of $X$ then $X$ must be accepted as a pseudo label. 
Rule \eqref{ind5} states that if the prediction of $X$ might be incorrect then $X$ could be accepted or not accepted as a pseudo label. Rule~\eqref{ind7} says that if $X$ is not selected as a pseudo label then $nok(X)$ is true.

\smallskip \noindent    
In summary, $\Pi$ is the collection of rules of the Rules~\eqref{over1}-\eqref{ind7}. Together with the set $A_S$ of the predicted labels and the knowledge base $KB_D$, we can compute a new set of pseudo labels by computing the answer sets of $\pi(S) = A_S \cup KB_D \cup \Pi$. 
Each answer set $W$ of $\pi(S)$ consists of a set of atoms of the form $ok(X)$. We say that $L(W) = \{X \mid ok(X) \in W\}$ is the set of pseudo labels corresponding to $W$. Intuitively, $L(W)$ encodes a revision of $A_S$. For an arbitrary sentence $S$ in a domain $D$, we can prove several properties of the program $\pi(S)$ such as 
(\emph{i}) $\pi(S)$ is consistent; 
(\emph{ii}) if $A_S$ does not contain any relation then $relation\_exists$ does not belong to any answer set of $\pi(S)$; and 
(\emph{iii}) for every $W$, $L(W)$ does not contain any overlapping pseudo labels if $overlap\_fl$ is true; and $L(W)$ does not contain type-inconsistent relations or entities.
Intuitively, (\emph{i}) represents the fact that there is always 
some revision of $A_S$. This can be proven using the splitting theorem proposed by \citeN{LifschitzT94}; (\emph{ii}) holds because of Rule~\eqref{re1}; and 
(\emph{iii}) holds due to the rules and constraints defined for type inference and checking. Due to space limitation, we omit the formal proof. 
The next example illustrates the use of $\Pi$. 
\begin{example}
    Consider the sentence in Example~\ref{eg:pl}. The program produces twenty answer sets (the $KB_{\tt CoNLL04}$ and the answer sets of the program are given in the Appendix). We observe that 
\begin{list}{$\bullet$}{\itemsep=0pt \parsep=0pt \topsep=0pt \leftmargin=12pt}
        \item Some answer set does not contain any atom of the form $ok(relation(\ldots))$; 
        \item No answer set contains both $ok(entity(org,0,2))$ and $ok(entity(other,1,2))$ because they overlap each other and $KB_{\tt CoNLL04}$ contains $overlap\_fl$; 
        \item $ok(relation(orgBasedIn,1,2,12,13))$ belongs to some answer sets but not all. It is because $entity(other,1,2)$ is of the type $other$ that does not match the required type ($org$) in the definition of $orgbasedIn$; and 
        \item     If $ok(relation(locatedIn,7,9,10,11))$ and $ok(relation(locatedIn,10,11,12,13))$ belong to an answer set then  $ok(relation(locatedIn,7,9,12,13))$ also belongs to the answer set  due to the transitivity of $locatedIn$, a part of the $KB_{\tt CoNLL04}$.
\end{list}
\end{example}

Note that encoding knowledge in ASP incurs extra works. However, compared with manually labeling a large amount of data, this extra works pay off.

\vspace{-0.1in}
\subsubsection{Computing the Most Preferred Answer Set}
\label{sec:choose_answer_set}

\begin{definition}[Preference score 
of an answer set]
\label{def:answerset_prob}
Given an answer set $W$, its preference score is defined as  
\vspace{-0.15in}
\begin{equation} \label{def:prob}  
        pref(W) = \Pi_{prob(a, p) \in W } p * \Pi_{invprob(a, p)   \in W} (1-p) 
    \end{equation}
\end{definition}
\noindent The preference score $pref(W)$ is the product of two terms, the first term is the product of the confidence level $p$ of every pseudo label $a$ such that $ok(a) \in W$ (hence, $prob(a,p) \in W$, due to Rule~\eqref{ind2}) and the second term is the product of the complement confidence level $1-p$ of pseudo label $a$ such that $ok(a) \not\in W$ (hence, $invprob(a,p) \in W$, due to Rule~\eqref{ind3}).
    It is easy to see that for two answer sets $A$ and $B$ such that $L(B) \subseteq L(A)$ (i.e., $A$ contains all acceptable labels in $B$), 
    if $prob(l, p) \in A $ and $p \ge 0.5$ for every $l \in L(A) \setminus L(B)$, then $p(A) \ge p(B)$. 
    Intuitively, this probability definition favors answer sets containing more (w.r.t. $\subseteq$) pseudo labels whose confidence level is greater than 0.5.    
When  $relation\_fl$ is set to true, we set $pref(W)=0$ 
if $relation\_exists \not\in W$. 
Preference will be given to the answer set with maximal probability. When all answer sets have zero preference, we prefer those with higher number of entity pseudo labels. The selection of the most preferred answer set is implemented using \lstinline{clingo} and its PythonAPI library.
The confidence level of an answer set $W$ (i.e., $W.conf$) is defined 
by $\min \{p \mid prob(l, p) \in W\}$. 

\vspace{-0.15in}
\section{Experiments}
\label{sec:exp}
The algorithms and models are implemented in Python 3.8 and run on a server with two 32-core GPU \verb|@|3.9 GHz, 512 GB RAM, and one NVIDIA A100 GPU. The Clingo version is 5.5.2. 

\noindent{\bf Data}. 
We use two datasets, CoNLL04 \cite{SPERT2020,DBLP:conf/conll/RothY04,DBLP:conf/coling/GuptaSA16,wang2020two} and SciERC \cite{SPERT2020,DBLP:conf/emnlp/LuanHOH18}, which have been utilized in other entity/relation extraction work. 
The CoNLL04 dataset contains sentences extracted from newspapers. We employ the training (922 sentences), development (231 sentences) and test set (288 sentences) split , which is similar to that of~\cite{DBLP:conf/coling/GuptaSA16}. 
The SciERC dataset consists of abstracts of artificial intelligence research papers. The training/development/test split is 1861/275/551 sentences, which is the same as that of \citeN{DBLP:conf/emnlp/LuanHOH18}. 

These datasets all have training sets. To utilize these datasets to verify the effectiveness of our method, we do not use  the whole training set to train the initial model. Instead, we use a small percentage of the training data (e.g., $p_{tr}\!\in\!(0,1]$) as the actual training data $D_L$ and use the remaining ($1-p_{tr}$) training data as the unlabeled data $D_{UL}$ to calculate pseudo labels. The original testing data is still utilized to  test the model performance. 
To get {\bf stable} results, for each dataset, we randomly choose five subsets (each one contains $p_{tr}$ of the training data) from the training data and train five models. Then, we report the averaged results from the five models. \\
{\bf Performance metric}. We report the micro and macro $F_1$ values for entities (E), relations (R), and relations together with entities (ER). 
The micro $F_1$ is calculated globally by counting the total true positives (TP), false negatives (FN), and false positives (FP).  In all the counting, a special class (not-an-entity or not-a-relation), which is encoded as zero, is never treated as positive.   
For example, considering the $E$ type, all the correctly predicted non-zero entities are counted as TP. Among the wrongly predicted entities,  an entity is counted as FP if it is wrongly predicted to be non-zero,  and an entity is counted as FN if its true class is non-zero. Some wrongly predicted entities are counted in both FP and FN.
The macro $F_1$ is obtained by calculating the prediction $F_1$ when treating each class as positive (and all the others as negative) and averaging the $F_1$s for all the classes.
We also report the running time of the models.  \\
\noindent {\bf Methods}. We compare our \METHOD~method with three classical and state-of-the-art baselines listed below.  
(1) {\em Self-training} as described in the papers by~\citeN{self-training2006}; \citeN{self-training2007}. 
For this method, we use 90\% as the threshold to select pseudo labels to be included for model retraining.
(2) {\em Curriculum-labeling (CL)}: this method retrains the model using the curriculum-labeling strategy proposed by \citeN{curriculum_labeling_aaai2021}. This is a state-of-the-art approach for SSL using pseudo labels. It has one hyper parameter (stepping threshold) controlling the confidence value of pseudo labels that are included in the training of the model in one iteration. This parameter is set to the same (20\%) as the original paper.  
(3) {\em Tri-training}: this method retrains the model using the tri-training strategy proposed by~\citeN{tri-training2005}. 
A recent study by \citeN{Ruder2018} has shown that the classic tri-training method is still a strong baseline for neural semi-supervised learning for natural language processing. 

For fair comparison, we run five iterations (retraining of the model) for every model. For our model, $\Delta$ is set to be 20\% as that in curriculum-labeling approach.

\vspace{-0.1in}
\subsection{Effectiveness Analysis}
We conduct experiments to examine the effectiveness of our approach. 
Our {\bf first set of experiments} is to compare our \METHOD~method with the other baselines. In these experiment, $p_{tr}$ is set to be 10\%. I.e., 10\% of the training data forms $D_L$.
Table~\ref{tb:method_comparison} shows the results on the CoNLL04  and the SciERC datasets. It shows that \METHOD~outperforms all the other baselines on all the calculated $F_1$ measurement on recognizing relations (R)  and both entities and relations (ER) no matter it is at the micro or macro level. For Entity (E) recognition, Tri-training is slightly better than our method. This is because our training process gives higher preferences to sentences with potentially correct relations. These results show the superiority of our proposed method. 

When more training data is available and the KB cannot provide extra information than what the labeled data can provide, ASPER may not beat the pure data driven models such as trile-training and curriculum labeling. However, ASPER is able to improve (at least not hurt) its base deep learning model (the SpERT model in this paper) that ASPER is built upon no matter whether the KB can provide much more information than the training data or not. 

\begin{table}[htb]
\centering
\begin{tabular}
{|c|c|c|c|c|c|c|c|}
\cline{1-7}
& \multicolumn{6} {@{}c@{}|}{CoNLL04 dataset} \\
  \cline{2-7}
& \multicolumn{3} {@{}c@{}|}{$F_1$ (micro)} 
  & \multicolumn{3} {@{}c@{}|} { $F_1$ (macro)}\\ 
  \cline{2-7}
 method & E & R & ER & E & R & ER\\
 \cline{1-7}
Self-train	
& 77.74$\pm$1.7 &	41.76$\pm$5.7 &	41.39$\pm$5.7 &	72.50$\pm$1.9 &	43.19$\pm$6.0 &	42.82$\pm$6.0 \\
\cline{1-7}
CL
&77.49$\pm$1.1 &	41.61$\pm$3.0 &	41.35$\pm$3.2 &	72.03$\pm$1.6 &	43.07$\pm$3.8 &	42.77$\pm$4.0\\
\cline{1-7}
Tri-train 
&78.63$\pm$2.4	& 42.60$\pm$6.7	& 42.29$\pm$6.7	& 72.49$\pm$2.5 &	42.99$\pm$7.1 & 42.64$\pm$7.2\\
\cline{1-7}
\METHOD
&{\bf 81.25}$\pm$1.2	&{\bf 52.47}$\pm$3.6	&{\bf 52.41}$\pm$3.6	&{\bf 75.90}$\pm$1.7	&{\bf 53.32}$\pm$4.0	&{\bf 53.27}$\pm$4.0
\\
\cline{1-7}
\end{tabular}

\begin{tabular}
{|c|c|c|c|c|c|c|c|}
\cline{1-7}
& \multicolumn{6} {@{}c@{}|}{SciERC dataset} \\
  \cline{2-7}
& \multicolumn{3} {@{}c@{}|}{$F_1$ (micro)} 
  & \multicolumn{3} {@{}c@{}|} { $F_1$ (macro)}\\ 
  \cline{2-7}
 method & E & R & ER & E & R & ER\\
 \cline{1-7}
Self-train	
& 56.72$\pm$1.2 &  18.60$\pm$2.6 &  12.36$\pm$1.7 &  54.43$\pm$1.4 &	11.07$\pm$3.7 &	6.98$\pm$2.3\\
\cline{1-7}
CL
& 60.75$\pm$0.8	&31.00$\pm$2.1 & 20.81$\pm$1.0 &59.19$\pm$0.4 & 22.00$\pm$3.8 & 15.55$\pm$1.8 \\
\cline{1-7}
Tri-train 
&{\bf 60.99}$\pm$0.7& 27.43$\pm$1.9 &	18.94$\pm$1.4 &	{\bf 59.52}$\pm$0.4 &	17.09$\pm$3.6 &	11.59$\pm$2.7\\
\cline{1-7}
\METHOD
&60.34$\pm$0.6 &{\bf 32.30}$\pm$1.2 &	{\bf 21.73}$\pm$1.2 & 59.10$\pm$0.4 &	{\bf 22.72}$\pm$3.1 &{\bf 16.06}$\pm$2.3\\
\cline{1-7}
\end{tabular}
\caption{Performance comparison of \METHOD~and other baselines on the two datasets ($E$: entity, $R$: relation, $ER$: entity and relation; $p_{tr}=10\%$)
\label{tb:method_comparison}}
\end{table}

We conduct a more detailed analysis about the running of \METHOD~by showing its performance in three iterations (other iterations show similar trend). 
This analysis is to show the quality of the pseudo label revision. The quality can be measured by comparing the pseudo labels and the ground truth labels (which are the training data with the correct labels, but are not directly used for training) and calculating the $F_1$ score.

\begin{table}[htb]
\centering
\begin{tabular}{|c|c|c|c|c|c|c|c|}
\cline{1-8}
\multicolumn{2}{|c|}{} & \multicolumn{6}{c|}{CoNLL04 dataset}\\
\cline{3-8}
 \multicolumn{2} {|c|}{}
  & \multicolumn{3} {c|} {$F_1$ (micro)} 
  & \multicolumn{3} {c|} {$F_1$ (macro)}\\ 
  \cline{3-8}
  \multicolumn{2} {|c|}{\METHOD} & E & R & ER & E & R & ER\\
 \cline{1-8}
Iter 1	
& no-ASP & 75.65 & 41.62 & 41.07 & 69.81 & 42.86 & 42.29\\\cline{2-8}
& with ASP & 77.06 & 44.45 & 44.45 & 71.16 & 45.40 & 45.40\\\cline{1-8}
Iter 2
& no-ASP & 78.68 &	49.78 &	49.08 &	73.24 &	50.86 &	50.16\\\cline{2-8}
& with ASP &79.01 &	50.19 & 50.19 & 73.49 &	51.27 & 51.27\\\cline{1-8}
Iter 3	
& no-ASP & 79.98 &	52.92 &	52.66 &	74.25 & 53.97 & 53.72\\\cline{2-8}
& with ASP & 80.24 & 53.62 & 53.62 & 74.52 & 54.67 & 54.67\\\cline{1-8}
\end{tabular}

\begin{tabular}{|c|c|c|c|c|c|c|c|}
\cline{1-8}
\multicolumn{2}{|c|}{} & \multicolumn{6}{c|}{SciERC dataset}\\
\cline{3-8}
 \multicolumn{2} {|c|}{}
  & \multicolumn{3} {c|} {$F_1$ (micro)} 
  & \multicolumn{3} {c|} {$F_1$ (macro)}\\ 
  \cline{3-8}
  \multicolumn{2} {|c|}{\METHOD} & E & R & ER & E & R & ER\\
 \cline{1-8}
Iter 1	
& no-ASP &  60.34 & 30.87 & 21.98 & 58.78 & 21.47 & 16.59\\\cline{2-8}
& with ASP &  60.34 & 31.12 & 22.09 & 58.78 & 21.75 & 16.79\\\cline{1-8}
Iter 2
& no-ASP & 60.86 &	32.59 & 23.10 &	59.39 & 22.34 & 17.25\\\cline{2-8}
& with ASP &60.86 & 32.83 & 23.24 & 59.39 &	22.58 & 17.42\\\cline{1-8}
Iter 3	
& no-ASP & 60.65 &	33.11 &	23.53 &	59.31 & 22.79 & 17.74\\\cline{2-8}
& with ASP & 60.65 & 33.12 & 23.54 & 59.31 & 22.82 & 17.76\\\cline{1-8}
\end{tabular}

\caption{The effect of the \METHOD~ algorithm to improve the quality of the pseudo labels on the two datasets. ($E$: entity, $R$: relation, $ER$: entity and relation; $p_{tr}=10\%$)
\label{tb:ASPER_iter_effect}}
\end{table}

Table~\ref{tb:ASPER_iter_effect} shows the detailed analysis of how the ASP component helps improve the quality of the generated pseudo labels. 
We can see that after each iteration, the F1 of the ASP generated pseudo labels is always higher than that of the raw pseudo labels. This confirms that the use of answer sets and ASP helps improve the quality of the pseudo labels. 
Table~\ref{tb:ASPER_iter_effect} also shows that the performance improvement in earlier iterations is better than that in later iterations. 
This is also consistent with our design of utilizing curriculum labeling, i.e., answer sets with higher confidence values are used in earlier iterations (Section~\ref{sec:model_retrain}).

The {\bf second} set of experiments examines the effect of the amount of initial training data on \METHOD. We change the amount of initial training dataset $D_L$ by varying the percentage $p_{tr}$ with different values (5\%, 10\%, 20\%, 30\%). 

\begin{figure}[htb]
    \begin{tabular}{cc}
    \includegraphics[width={.4\columnwidth}]{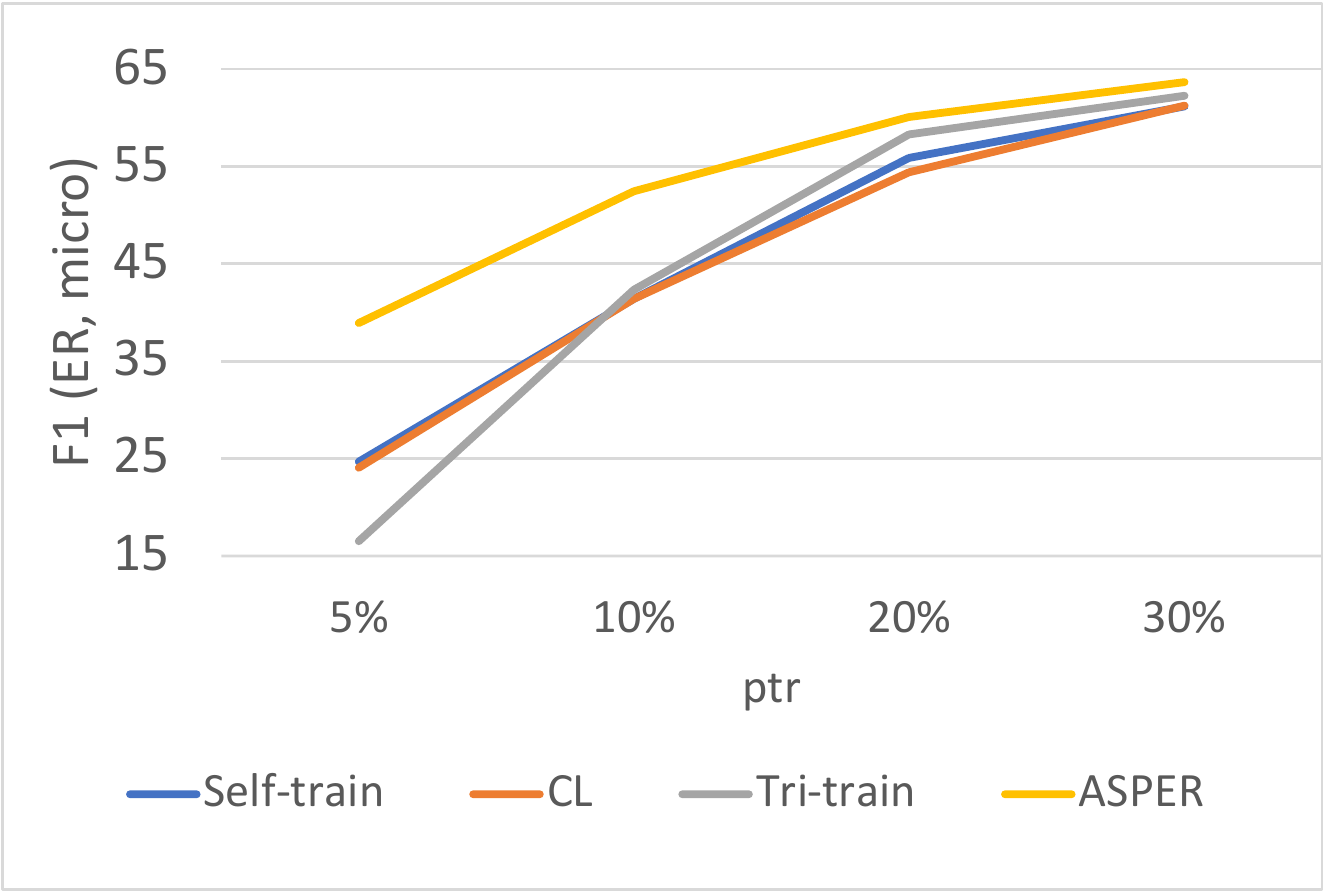}&
    \includegraphics[width={.4\columnwidth}]{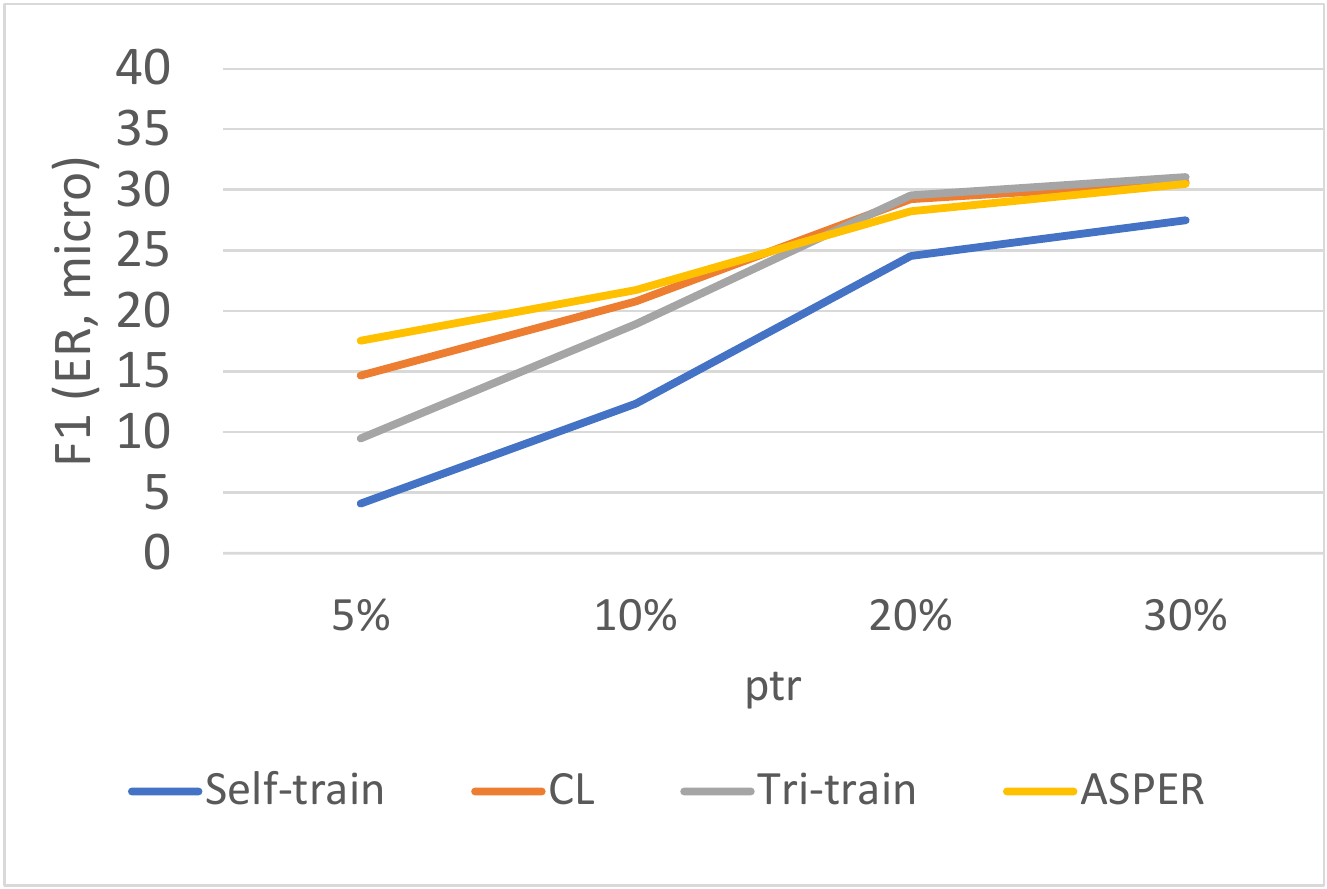}\\
    (a) CoNLL04 & (b) SciERC
    \end{tabular}
    \caption{Performance comparison (varying training data amount)}
    \label{fig:ASPER_vary_ptr}
\end{figure}

The results are reported in Figure~\ref{fig:ASPER_vary_ptr}. We only report the F1 of relation with entity (micro) because the trend of the other F1 values is the same. The figure shows that, when there is less initial training data, the overall performance is worse. However the positive effect of \METHOD~is more obvious when there are less training data,   which can be observed from the larger gap between \METHOD~and other methods for smaller $p_{tr}$ (e.g., 5\%). This result is consistent with our intuition of designing \METHOD~to alleviate the issue of insufficient amount of training data.  Figure~\ref{fig:ASPER_vary_ptr}(b) also shows that our method does outperform, but has comparable performance as, CL and tri-training when $p_{tr}$ is larger. The major reason is that the knowledge base is less  effective in capturing the characteristics of the second domain (research articles). More effective rules need to be developed to enrich the knowledge base in the future.

The third set of experiments conduct an {\bf ablation study} to investigate the effect of the rules in the ASP program. 
Due to space limitation, we present this analysis on one dataset. Table~\ref{tb:ablation_study} shows the results. 
\begin{table}[htb]
\centering
\begin{tabular}{|l|c|c|c|c|c|c|} 
\cline{1-7}
& \multicolumn{3} {c|} {($F_1$ micro)} 
  & \multicolumn{3} {c|} { $F_1$ (macro)} \\ 
\cline{2-7}  
  & E & R & ER & E & R & ER\\ 
\cline{1-7}
with all rules &{\bf 81.25} &	{\bf 52.47} & {\bf 52.41} & {\bf 75.90}& {\bf 53.32}& {\bf 53.27}\\
\cline{1-7}
with all rules except the $relation\_exists$ rule
&76.64	& 34.13	& 33.84	& 70.56	& 34.87	& 34.57\\ 
\cline{1-7} 
without any rules
&76.74	& 31.07	& 31.07	& 70.52	& 31.99	& 31.99\\ 
\cline{1-7} 
\end{tabular}
\caption{Ablation study on \METHOD; 
CoNLL04} 
\label{tb:ablation_study}
\end{table}
The first row ({\em with all rules}) shows the results when all the rules are utilized. The third row ({\em without any rules}) on the other hand shows the results when no rule is utilized. The results (improvement of the first row comparing to the third row) clearly demonstrate that the rules contribute positively to the performance of \METHOD. 
We conduct a further analysis about the effect of the different type of rules and find that the $relation\_exists$ rule (Rule~\eqref{re1}) plays the most significant role. The second row shows the results from the program while the $relation\_exists$ rule is not utilized, but all the other rules are used. The improvement of all the other rules to the algorithm (which is captured by the difference between the 2nd and the 3rd row) is not as much as the $relation\_exists$ rule (which is observed from the difference between the 1st and the 2nd row).

\vspace{-0.2in}
\subsection{Efficiency Analysis}
We also examine the running time of the different methods to understand the overhead brought by the ASP program. Due to space constraint, we report the summarized data here. 
Self training, curriculum labeling, and \METHOD~use similar amount of time. On the CoNLL04 dataset, it takes approximately 40-50 minutes to run the five iterations. Tri-training's time is approximately three times of the other three methods because it needs to train three models in each iteration. The overhead of using ASP to generate the updated pseudo labels is about 30 seconds in each iteration. This time is negligible compared with the expensive NN model training. 

\vspace{-0.05in}
\section{Conclusions}
\label{sec:conclusion}
In this paper, we presented a novel method \METHOD, which leverages Answer Set Programming (ASP) to improve the performance of Neural Network models in the joint recognition of entities and relations from text data when limited amount of training data is available. 
\METHOD~makes use of pseudo labels. 
The ASP program encodes different types of commonsense rules by taking advantage of the commonsense domain knowledge.
The experiments on two real datasets show that \METHOD~can report significantly better results than the other baselines in most cases.

\bibliographystyle{acmtrans}
\bibliography{reference}
\section{Appendix}
\subsection{Encoding and Output of Revising Pseudo Labels of Example ~\ref{eg:pl}}
The collection $A_S$ for the sentence in Example~\ref{eg:pl}:
{\small
\begin{verbatim}
atom(entity(org,0,2),"0.888"). % CDT Tuesday
atom(entity(other,1,2),"0.799"). % Tuesday
atom(entity(loc,7,9),"0.998"). % Port Ather
atom(entity(loc,10,11),"0.998"). % Galveston
atom(entity(loc,12,13),"0.998"). % Texas
atom(relation(locatedIn,7,9,10,11),"0.998"). % locatedIn(Port Ather,Galveston)}
atom(relation(locatedIn,10,11,12,13),"0.993").% locatedIn(Galveston,Texas)
atom(relation(locatedIn,0,2,12,13),"0.993").% locatedIn(CDT Tuesday,Texas)
atom(relation(orgbasedIn,1,2,12,13),"0.777").   % locatedIn(Tuesday,Texas)
\end{verbatim}
}

\smallskip 
\noindent 
The domain dependent $KB_D$ for the \lstinline{ConLL04} dataset:

{\small
\begin{verbatim}
type_def(liveIn, peop, loc).
type_def(locatedIn, loc, loc).
type_def(orgbasedIn, org, loc).
type_def(workFor, peop, org).
type_def(kill, peop, peop).

rule(relation(locatedIn,P1,P2,Q1,Q2), 
     relation(orgbasedIn,O1,O2,P1,P2), 
     relation(orgbasedIn,O1,O2,Q1,Q2)):-
  atom(relation(locatedIn,P1,P2,Q1,Q2)), 
  atom(relation(orgbasedIn,O1,O2,P1,P2)), 
  not atom(relation(orgbasedIn,O1,O2,Q1,Q2)),
  P1 != P2, P1 != Q1.

rule(relation(locatedIn,P1,P2,Q1,Q2), 
     relation(locatedIn,Q1,Q2,R1,R2), 
     relation(locatedIn,P1,P2,R1,R2)):-
  atom(relation(locatedIn,P1,P2,Q1,Q2)), 
  atom(relation(locatedIn,Q1,Q2,R1,R2)), 
  not atom(relation(locatedIn,P1,P2,R1,R2)),
  P1 != P2, P1 != Q1, Q1 != Q2, Q1 != R1.

rule(relation(liveIn,X1,X2,P1,P2), 
     relation(locatedIn,P1,P2,Q1,Q2), 
     relation(liveIn,X1,X2,Q1,Q2)):-
  atom(relation(liveIn,X1,X2,P1,P2)), 
  atom(relation(locatedIn,P1,P2,Q1,Q2)), 
  not atom(relation(liveIn,X1,X2,Q1,Q2)),
  P1 != P2, P1 != Q1.

overlap_flag.
relation flag.
\end{verbatim}

\smallskip 
\noindent 
The twenty answer sets of the program $\pi(S)$ for the sentence from Example~\ref{eg:pl} (the listing contains only atoms of the form $ok(.)$ each represents a pseudo label):
{\small
\begin{verbatim}
Answer: 1 (pref=0, conf=0)
ok(entity(loc,7,9)) ok(entity(loc,10,11))  ok(entity(loc,12,13))
Answer: 2 (pref=0, conf=0)
ok(entity(loc,7,9)) ok(entity(loc,10,11)) ok(entity(loc,12,13)) 
ok(entity(org,0,2))
Answer: 3 (pref=0, conf=0)
ok(entity(loc,7,9)) ok(entity(loc,10,11)) ok(entity(loc,12,13)) 
ok(entity(other,1,2))
Answer: 4 (pref=1.7039341161594381e-09, conf=0.777)
ok(entity(loc,7,9)) ok(entity(loc,10,11)) ok(entity(loc,12,13)) 
ok(entity(loc,0,2)) ok(relation(locatedIn,0,2,12,13))
Answer: 5 (pref=6.937258075900524e-08, conf=0.993)
ok(entity(loc,7,9)) ok(entity(loc,10,11))  ok(entity(loc,12,13)) 
ok(entity(org,1,2)) ok(relation(orgbasedIn,1,2,12,13))
Answer: 6 (pref=6.937258075900524e-08, conf=0.993)
ok(entity(loc,7,9)) ok(entity(loc,10,11)) ok(entity(loc,12,13)) 
ok(relation(locatedIn,10,11,12,13))
Answer: 7 (pref=5.50025461732113e-07, conf=0.888)
ok(entity(loc,7,9)) ok(entity(loc,10,11)) ok(entity(loc,12,13)) 
ok(entity(other,1,2)) ok(relation(locatedIn,10,11,12,13))
Answer: 8 (pref=2.757646369474885e-07, conf=0.799)
ok(entity(loc,7,9)) ok(entity(loc,10,11)) ok(entity(loc,12,13)) 
ok(entity(org,0,2)) ok(relation(locatedIn,10,11,12,13))
Answer: 9 (pref=9.840996099098876e-06, conf=0.993)
ok(entity(loc,7,9)) ok(entity(loc,10,11)) ok(entity(loc,12,13)) 
ok(entity(org,1,2)) ok(relation(locatedIn,10,11,12,13)) 
ok(relation(orgbasedIn,1,2,12,13))
Answer: 10 (pref=2.4171522533518863e-07, conf=0.777)
ok(entity(loc,7,9)) ok(entity(loc,10,11)) ok(entity(loc,12,13)) 
ok(relation(locatedIn,10,11,12,13)) ok(relation(locatedIn,0,2,12,13))
ok(entity(loc,0,2)) 
Answer: 11 (pref=2.4402661086727617e-07, conf=0.998)
ok(entity(loc,7,9)) ok(entity(loc,10,11)) ok(entity(loc,12,13)) 
ok(relation(locatedIn,7,9,10,11))
Answer: 12 (pref=9.700361297659388e-07, conf=0.799)
ok(entity(loc,7,9)) ok(entity(loc,10,11)) ok(entity(loc,12,13)) 
ok(entity(org,0,2)) ok(relation(locatedIn,7,9,10,11))
Answer: 13 (pref=1.934782414733404e-06, conf=0.888)
ok(entity(loc,7,9)) ok(entity(loc,10,11)) ok(entity(loc,12,13)) 
ok(entity(other,1,2)) ok(relation(locatedIn,7,9,10,11))
Answer: 14 (pref=3.4616917798743576e-05, conf=0.993)
ok(entity(loc,7,9)) ok(entity(loc,10,11)) ok(entity(loc,12,13)) 
ok(relation(locatedIn,7,9,10,11)) ok(relation(locatedIn,0,2,12,13))
ok(entity(loc,0,2))
Answer: 15 (pref=8.502631239635589e-07, conf=0.777)
ok(entity(loc,7,9)) ok(entity(loc,10,11)) ok(entity(loc,12,13)) 
ok(relation(locatedIn,7,9,10,11)) ok(relation(orgbasedIn,1,2,12,13))
ok(entity(org,1,2))
Answer: 16 (pref=3.4616917798743576e-05, conf=0.993)
ok(entity(loc,7,9)) ok(entity(loc,10,11)) ok(entity(loc,12,13)) 
ok(relation(locatedIn,7,9,10,11)) ok(relation(locatedIn,10,11,12,13)) 
ok(relation(locatedIn,7,9,12,13))
Answer: 17 (pref=0.0002744627054043241, conf=0.888)
ok(entity(loc,7,9)) ok(entity(loc,10,11)) ok(entity(loc,12,13)) 
ok(entity(other,1,2)) ok(relation(locatedIn,7,9,10,11)) 
ok(relation(locatedIn,10,11,12,13)) ok(relation(locatedIn,7,9,12,13))
Answer: 18 (pref=0.00013760655383679662, conf=0.799)
ok(entity(loc,7,9)) ok(entity(loc,10,11)) ok(entity(loc,12,13)) 
ok(entity(org,0,2)) ok(relation(locatedIn,7,9,10,11)) 
ok(relation(locatedIn,10,11,12,13)) ok(relation(locatedIn,7,9,12,13))
Answer: 19 (pref=0.004910657053450334 (maximum prob), conf=0.993)
ok(entity(loc,7,9)) ok(entity(loc,10,11)) ok(entity(loc,12,13)) 
ok(relation(locatedIn,7,9,10,11)) ok(relation(locatedIn,10,11,12,13)) 
ok(relation(locatedIn,0,2,12,13)) ok(relation(locatedIn,7,9,12,13)) 
ok(entity(loc,0,2))
Answer: 20 (pref=0.00012061589744225903, conf=0.777)
ok(entity(loc,7,9)) ok(entity(loc,10,11)) ok(entity(loc,12,13)) 
ok(relation(locatedIn,7,9,10,11)) ok(relation(locatedIn,10,11,12,13)) 
ok(relation(orgbasedIn,1,2,12,13)) ok(relation(locatedIn,7,9,12,13)) 
ok(entity(org,1,2)) 
\end{verbatim}
}
\end{document}